# A NEURAL DOCUMENT LANGUAGE MODELING FRAMEWORK FOR SPOKEN DOCUMENT RETRIEVAL

*Li-Phen Yen, Zhen-Yu Wu, and Kuan-Yu Chen*

National Taiwan University of Science and Technology

## ABSTRACT

Recent developments in deep learning have led to a significant innovation in various classic and practical subjects, including speech recognition, computer vision, question answering, information retrieval and so on. In the context of natural language processing (NLP), language representations learned by referring to autoregressive language modeling or autoencoding have shown giant successes in many downstream tasks, so the school of studies have become a major stream of research recently. Because the immenseness of multimedia data along with speech have spread around the world in our daily life, spoken document retrieval (SDR), which aims at retrieving relevant multimedia contents to satisfy users' queries, has become an important research subject in the past decades. Targeting on enhancing the SDR performance, the paper concentrates on proposing a neural retrieval framework, which assembles the merits of using language modeling (LM) mechanism in SDR and leveraging the abstractive information learned by the language representation models. Consequently, to our knowledge, this is a pioneer study on supervised training of a neural LM-based SDR framework, especially combined with the pretrained language representation methods. A series of empirical SDR experiments conducted on a benchmark collection demonstrate the good efficacy of the proposed framework, compared to several existing strong baseline systems.

## 1. INTRODUCTION

Spoken content analysis has become an attractive and rising research subject over the past two decades in the speech processing community, because large volumes of multimedia data associated with spoken documents made available to the public [1-3]. There are two main streams of research on processing a given text/spoken query and a spoken document. On one hand, spoken term detection [4, 5] embraces the goal of extracting probable spoken terms or phrases inherent in a spoken document that could match the query words or phrases literally. On the other hand, spoken document retrieval (SDR) [4, 6] revolves more around the notion of relevance of a spoken document in response to the query. It is generally agreed upon that a document is relevant if it could address the stated information need of the query, not because it just happens to contain all the words in the given query [7, 8]. Within the research of SDR and conventional information retrieval (IR), a significant body of studies has been devoted to developing effective retrieval methods, such as the vector space model [17], the Okapi BM25 model [18], the topic models [21-23], and so forth. Among them, recently, the emerging paradigm turns to employ a statistical language modeling (LM) approach for SDR [19, 20]. This paradigm attracts much interest due to its simplicity and clear probabilistic meaning, as well as state-of-the-art performance. Accordingly, a given text or spoken document can be framed as a generative document language model for observing a query, while the query is regarded as an observation, expressed by a sequence of words. Therefore, documents can be ranked according to their likelihoods of generating the query, namely the query-likelihood measure (QLM) [7, 8, 19]. In order to further boost the performance of using LM in SDR (and IR), several document language modeling methods have been introduced, including the probabilistic topic models [21-23], the positional language models [24], the parsimonious language models [25], to name but a few.

On a separate front, deep learning has gained significant interest of research and experimentation in many applications because of its remarkable performance [9, 10, 11]. When it comes to the research field of natural language processing (NLP), language representation methods can be viewed as pioneering studies [13-15, 26-30]. Previous research developed in this vein includes the CBOW [14], the skipgram [14], the GloVe [15], and so on. The central idea of these methods is to learn continuously distributed vector representations of words using neural networks, which can probe latent semantic and/or syntactic cues that can in turn be used to induce similarity measures among words. The line of research recently goes to manipulate large and deep neural network architectures, which are usually composed of a stack of the transformer blocks [12]. The major innovation of these models is to capture the interactions between each pair of input tokens by using the bidirectional self-attention mechanism [11, 12, 28] and to update the model parameters by optimizing the masked language modeling objective [28], which can alleviate the constraint of conventional left-to-right autoregressive objective [26, 27]. Moreover, these models are designed to pretrain deep bidirectional language representations, so they can be adapted to various downstream tasks by simply plugging in the minimal task-specific parameters, which will be fine-tuned in addition to the pretrained language representations for the downstream task. Representative methods include, but are not limited to, the OpenAI GPT [27], the BERT [28], the XLNET [29], the RoBERTa [30], and the ALBERT [31].

Motivated by these observations, the paper strives to develop an efficient and effective SDR framework, which targets at assembling the merits of using language modeling mechanism in SDR and leveraging the abstractive information learned by the language representation methods. As far as we are aware, this is a pioneer research on supervised training of a neural LM-based SDR framework, especially combined with the novel language representation methods. To sum up, the major contributions of this paper are at least three-fold. First, a novel neural LM-based SDR framework, which is a conjunction of the LM mechanism and the language representation methods, is proposed. Second, stemming from such a framework, we make a step forward to introduce three different training objectives for inferring the model parameters.

Finally, a series of empirical evaluations and comparisons are conducted on a benchmark SDR corpus.

## 2. RELATED WORK

### 2.1. The Language Representation Methods

Because of the impressive successes in many NLP-related tasks, language representations have become a popular research recently. In general, the research spectrum can be classified into two main streams according to the usages for the downstream tasks [28]: (1) feature-based models and (2) fine-tuning methods. Famous and well-practiced representatives for the feature-based models are the word embedding methods. The neural network language model [13] is the most-known seminal study on developing various word embedding methods. It estimates a statistical ($N$-gram) language model, formalized as a feed-forward fully-connected neural network, for predicting future words while inducing word embeddings as a by-product. Such an attempt has already motivated many follow-up extensions to develop similar methods for probing latent semantic and syntactic regularities in the representation of words. Then, the learned word embeddings are usually treated as feature vectors for downstream tasks. Representative methods include, but are not limited to, the CBOW [14], the skipgram [14], the GloVe [15], and the ELMo [26]. On the contrary, the OpenAI GPT [27], the BERT [28], the XLNET [29], the RoBERTa [30], and the ALBERT [31] are the leading methods in the latter class. The fine-tuning methods usually consist of two parts: pretraining and task-specific parameters tuning. Formally, such a school of methods usually leverage an unsupervised objective to obtain a pretrained model, and then they introduce minimal task-specific parameters and train on the downstream task by simply fine-tuning all the (or only the task-specific) parameters [32-34]. Recently, the latter category (i.e., fine-tuning approaches) becomes preferable vehicle in NLP community.

### 2.2. Language Modeling for SDR

Owing to the fact that language modeling (LM) method has sound theoretical underpinnings and excellent empirical performance, it has become one of major mechanisms to build an SDR system [4, 19, 20]. The fundamental principle of using LM method to SDR is to determine the relevance degree between a pair of query $Q$ and spoken document $D$ by computing the conditional probability $P(Q|D)$, i.e., the likelihood of $Q$ generated by $D$ [19]. Such an approach is the so-called query-likelihood measure (QLM). A spoken document $D$ is deemed to be relevant to the query $Q$ if the corresponding document language model is more likely to generate the query. If the query $Q$ is treated as a sequence of words, $Q = q_1, q_2, \cdots, q_{|Q|}$, where the query words are assumed to be conditionally independent given the document $D$ and their order is also assumed to be of no importance (i.e., the so-called "*bag-of-words*" assumption), the similarity measure $P(Q|D)$ can be further decomposed as a product of the probabilities of the query words generated by the document [19, 20, 23]:

$$P(Q|D) = \prod_{w \in V} P(w|D)^{c(w,Q)} \tag{1}$$

where $V$ denotes the vocabulary, $c(w, Q)$ is the occurrence frequency of word $w$ in the query $Q$, $P(w|D)$ is the likelihood of generating word $w$ by document $D$, which is estimated based on the occurrence frequency of $w$ in $D$ by the maximum-likelihood estimator. To model the general properties of a language as well as to avoid the problem of zero probability, $P(w|D)$ is usually smoothed by a background unigram model $P(w|BG)$ [19, 20, 23].

Following the school of research, several document language modeling methods, which aim at inferring a more robust unigram probability $P(w|D)$, have been exploited, such as the PLSA [21], the LDA [22, 23], the positional language models [24], and the parsimonious language models [25], just to name but a few.

## 3. THE NEURAL DOCUMENT LANGUAGE MODELING FRAMEWORK

### 3.1 The Proposed Methodology

Recently, among the popular language representation methods, the bidirectional encoder representations from transformers (BERT) model [28] has attracted much interest due to its state-of-the-art performances in several NLP-related tasks. A naïve but efficient way is to treat BERT, which is a fine-tuning method, as an input encoder, and then a set of simple and task-specific layers is stacked upon the BERT model. Subsequently, the additional task-specific parameters can be fine-tuned toward to optimize the performance of the target task [32-34]. When BERT comes to the SDR (and IR) task, a straightforward strategy is to be employed to encode a concatenation word sequence of a query and a document, and then a simple classification objective is introduced to indicate whether the document is relevant to the query or not. More formally, for a query $Q = \{q_1, q_2, \ldots q_{|Q|}\}$ and a document $D = \{d_1, d_2, \ldots, d_{|D|}\}$, a concatenation word sequence $\{[CLS], q_1, q_2, \ldots q_{|Q|}, [SEP], d_1, d_2, \ldots, d_{|D|}, [SEP]\}$ can be obtained, where "[CLS]" represents a special token of every concatenation word sequence and "[SEP]" is a separator token. Next, the pretrained BERT model is used to extract a set of hidden vectors for each token in the concatenation word sequence. In order to infer a relevance degree between the query and the document, a single layer neural network is adopted to translate the hidden vector corresponding to the "[CLS]" token to the relevance score [34]. We term the model as "vanilla BERT". Although the naïve strategy can already obtain a certain level of performance, but it seems to limit the power of the BERT model. Accordingly, on top of the BERT, a neural language modeling-based framework is presented in this paper to further improve the performance of SDR.

#### 3.1.1 Direct Modeling Strategy

Our idea comes from the principle of using LM to SDR. For a given pair of query and document, the relevance degree can be determined by calculating the query likelihood $P(Q|D)$. Thus, a natural objective of a supervised retrieval model is to optimize the total log-likelihood of the training queries generated by their respective relevant documents. As an illustration, for a given set of training queries $\mathbf{Q} = \{Q_1, Q_2, \cdots, Q_{|\mathbf{Q}|}\}$ and the associated query-document relevance information $\mathbf{R} = \{R_{Q_1}, R_{Q_2}, \cdots, R_{Q_{|\mathbf{Q}|}}\}$, the training objective can be expressed by:

$$\operatorname*{argmax}_{\theta} \sum_{Q_i \in \mathbf{Q}} \sum_{D_j \in R_{Q_i}} \log P(Q_i|D_j) \\ = \sum_{Q_i \in \mathbf{Q}} \sum_{D_j \in R_{Q_i}} \sum_{w \in V} c(w, Q_i) \log P(w|D_j) \tag{2}$$

where $V$ denotes the vocabulary, $c(w, Q_i)$ is the occurrence count of word $w$ in query $Q_i$, and $R_{Q_i}$ is a set of relevant documents to training query $Q_i$. It is obvious that $P(w|D_j)$ plays a central role in the formulation, thus our goal is to design an enhanced document language modeling framework by neural networks. Accordingly, the proposed framework consists of two modules: a document encoder $f(\cdot)$, which can automatically infer a desired vector representation by encapsulating the entire information of a given document; and a language model generator $h(\cdot)$ that can produce a language model for an input document by considering its vector representation. To crystalize the idea to work, we implement the document encoder $f(\cdot)$ with the BERT model. More formally, for a document $D_j = \{d_1, d_2, ..., d_{|D_j|}\}$, the special tokens "[CLS]" and "[SEP]" are appended at the beginning and end of the document, i.e., $\{[CLS], d_1, d_2, ..., d_{|D_j|}, [SEP]\}$. Next, the pretrained BERT model is used to extract hidden vectors for each token in the sequence, and we treat the hidden vector for the "[CLS]" token as a comprehensive document representation, which can then be used to derived an enhanced language model for the document $D_j$ by the generator $h(\cdot)$. In this study, $h(\cdot)$ is a single layer neural network with parameter $\theta$, and the softmax activation function is applied on the output layer. Hence, the output of $h(\cdot)$ is a probability vector, where each element corresponds to the probability of a word/term of the vocabulary $V$. On a separate front, it is worthy to note that to maximize the total log-likelihood objective function (i.e., Eq(2)) is equivalent to minimize the cross entropy between the query language model and the document language model, when the query model is the simple maximum likelihood estimation. In a nutshell, to put everything together, our training objective comes to:

$$\begin{aligned}
&\underset{\theta}{\operatorname{argmax}} \sum_{Q_i \in \mathbf{Q}} \sum_{D_j \in R_{Q_i}} \sum_{w \in V} c(w, Q_i) \log P'(w|D_j) \\
&= \underset{\theta}{\operatorname{argmax}} \sum_{Q_i \in \mathbf{Q}} \sum_{D_j \in R_{Q_i}} \sum_{w \in V} \frac{c(w, Q_i)}{|Q_i|} \log P'(w|D_j) \\
&= \underset{\theta}{\operatorname{argmax}} \sum_{Q_i \in \mathbf{Q}} \sum_{D_j \in R_{Q_i}} \sum_{w \in V} P(w|Q_i) \log P'(w|D_j) \\
&= \underset{\theta}{\operatorname{argmin}} \sum_{Q_i \in \mathbf{Q}} \sum_{D_j \in R_{Q_i}} H\left(P_{Q_i} || P'_{D_j}\right) \\
&= \underset{\theta}{\operatorname{argmin}} \sum_{Q_i \in \mathbf{Q}} \sum_{D_j \in R_{Q_i}} H\left(P_{Q_i} || h\left(f(D_j)\right)\right)
\end{aligned} \quad (3)$$

where $H(\cdot || \cdot)$ denotes the function of cross entropy, $P_{Q_i}$ is the query language model, $P'_{D_j}$ (and $P'(w|D_j)$) represents the enhanced document language model, which is derived by the document encoder $f(\cdot)$ and the language model generator $h(\cdot)$. It should be noted that, in order to retain the valuable language information learned from a large collection of text data, the BERT model is pretrained[1] and fixed [28].

### 3.1.2 Pair-wise Modeling Strategy

Although the direct modeling (c.f. section 3.1.1) presents a novel and systematic way to employ BERT for inferring an enhanced document language model, such a method only takes positive examples into considerations. However, in the SDR task, the goal is to rank relevant documents higher than the irrelevant documents, thus a natural variant extended from the direct modeling strategy is a pair-wise modeling method. For a given set of training queries $\mathbf{Q} = \{Q_1, Q_2, \cdots, Q_{|\mathbf{Q}|}\}$ and the associated query-document relevance information $\mathbf{R} = \{R_{Q_1}, R_{Q_2}, \cdots, R_{Q_{|\mathbf{Q}|}}\}$, the network can be fine-tuned by referring to a pair-wise hinge loss function [35]:

$$\underset{\theta}{\operatorname{argmin}} \sum_{Q_i \in \mathbf{Q}} \sum_{D_j \in R_{Q_i}} \sum_{D_k \notin R_{Q_i}} max\left(0, 1 + H\left(P_{Q_i} || h\left(f(D_j)\right)\right) - H\left(P_{Q_i} || h(f(D_k))\right)\right) \quad (4)$$

where $P_{Q_i}$ is the query language model estimated by the maximum-likelihood estimator, $h\left(f(D_j)\right)$ and $h\left(f(D_k)\right)$ are enhanced document language models for relevant document $D_j$ and irrelevant document $D_k$, respectively. Based on the pair-wise hinge loss, the model not only tries to make a bridge between each training query and its relevant documents, but also distinguishes each pair of query and its irrelevant document.

---

[1] https://github.com/google-research/bert#pre-trained-models

### 3.1.3 Triple-wise Modeling Strategy

To take both relevant and irrelevant information into account, opposite to the pair-wise hinge loss, we further introduce a triple-wise training criterion in this paper. Formally, for each training step, we randomly select a training query $Q_i \in \mathbf{Q}$, one of its relevant documents $D_j \in R_{Q_i}$, and two different irrelevant documents $\{D_k, D_l\} \notin R_{Q_i}$. Subsequently, the triple-wise training objective function is defined by:

$$\underset{\theta}{\operatorname{argmax}} \sum_{Q_i \in \mathbf{Q}} \sum_{D_j \in R_{Q_i}} \frac{\exp^{-H\left(P_{Q_i} || h\left(f(D_j)\right)\right)}}{\exp^{-H\left(P_{Q_i} || h\left(f(D_j)\right)\right)} + \exp^{-H\left(P_{Q_i} || h\left(f(D_k)\right)\right)} + \exp^{-H\left(P_{Q_i} || h\left(f(D_l)\right)\right)}} \quad (5)$$

where $H(\cdot || \cdot)$ denotes the function of cross entropy. By bringing more irrelevant information at the training stage, we expect to increase the generalization ability of the proposed framework.

## 3.2 The Enhanced Document Language Model

On top of the proposed framework, the enhanced document language model can be inferred for each document. According to the literatures [7, 19, 20, 23], we all agree upon that the background language model can provide the general properties of a language and the original document information is still a crucial reference. Hence, the final document language model $\bar{P}(w|D)$ is a combination of the three necessary component models:

$$\bar{P}(w|D) = \alpha \cdot P(w|BG) + \beta \cdot P(w|D) + \gamma \cdot P'(w|D) \quad (6)$$

where $P(w|BG)$ denotes the background language model, $P(w|D)$ is the document language model estimated by the simple maximum likelihood estimator [19], $P'(w|D)$ (i.e., $h(f(D))$) is the enhanced document language model proposed in this paper, and $\alpha, \beta$ as well as $\gamma$ are empirical weighting factors. To sum up, the paper presents a pioneer study on supervised training of a neural LM-based SDR framework, which introduces a systematic way to combine a pretrained language representation method with the principle of using LM for SDR. Moreover, three training criterions are also investigated in this study for deriving the model parameters.

## 4. EXPERIMENTS

### 4.1 Experimental Setup

We used the Topic Detection and Tracking collection (TDT-2) [36] in the experiments. The Mandarin news stories from Voice of America news broadcasts were used as the spoken documents. All news stories were exhaustively tagged with event-based topic labels, which served as the relevance judgments for performance evaluation. The average word error rate obtained for the spoken documents is about 35% [37]. The Chinese news stories from Xinhua News Agency were used as our test queries. More specifically, in the following experiments, we will either use a whole news story as a "long query," or merely extract the tittle field from a news story as a "short query." To obtain the model parameters, 819 training query exemplars with the corresponding relevant documents are compiled. The retrieval performance is evaluated with the commonly-used non-interpolated mean average precision (MAP) following the TREC evaluation [38].

### 4.2 Experimental Results

At the beginning of the experiments, we classify the existing popular SDR systems into three categories: the unsupervised vector space-based models, the unsupervised LM-based methods, and the supervised neural retrieval models. In the first set of experiments, we explore the efficacies of the unsupervised vector space-based models, including the vector space model (VSM) [17], three classic word embedding methods (i.e., the CBOW [14], the skipgram [14], and the GloVe [15]), and three paragraph embedding methods (i.e., the DM [16], the DBOW [16], and the EV [39]) for SDR. Among these systems, each query and document is represented by a vector, and the relevance degree is computed by the cosine similarity measure. Furthermore, in this study, we also make a comparison between SDR and traditional text retrieval. Consequently, the retrieval results, assuming manual transcripts for the spoken documents to be retrieved (denoted by TD) are known, are also shown for reference, compared to the results when only the erroneous transcripts by speech recognition are available (denoted by SD). Experimental results are shown in the first block of Table 1. The best result within each column (corresponding to a specific evaluation condition) is type-set boldface. Inspection of these results reveals some noteworthy points. First, all the word embedding and paragraph embedding-based methods outperform conventional VSM by a large margin. Second, for paragraph embedding-based methods, EV and DBOW appear to be more flexible than DM, and EV usually achieves better results on the short query cases than DBOW while DBOW outperforms EV on the long query cases.

Next, in the second set of experiments, we further compare the unsupervised vector space-based models with the unsupervised LM-based methods, including the query-likelihood measure (QLM [19], cf. section 2.2) and the LDA [23]. The results are presented in the second block of Table 1. Notable observations can be summarized as follows. First, LDA outperforms QLM in all cases, thus the results confirm that deriving a more accurate document language model can really benefit the SDR performance. Second, language model-based methods in general outperform vector space-based methods. The results demonstrate that language model-based methods are a school of efficient and effective mechanism for SDR.

In the third set of experiments, we make a step forward to compare these unsupervised systems (i.e., vector space-based and LM-based methods) with the recently SOTAs (i.e., supervised neural retrieval models), including the GPR [40], the LPEV [41], the DSSM [42] and the vanilla BERT [34] (c.f. section 3.1). It is worthwhile to mention that the GPR, the LPEV and the DSSM aim at learning vector representations for queries and documents. Consequently, at query time, they use the cosine similarity measure to quantify the relevance degree between a query and a document with the learned embedding vectors. On the contrary, vanilla BERT directly maps the input concatenation word sequence to a relevance score. All the retrieval results are listed in the third part of Table 1. At the first glance, to our expected, the supervised neural retrieval models outperform other unsupervised baseline systems by a large margin. Among the neural retrieval models, GPR achieves the worst results than others, because it mainly focuses on mitigating the time-consuming problem instead of aiming at the SDR performance. We should also highlight that the simple vanilla BERT has obtained unneglected results on the SDR task, which demonstrates the capability and effectiveness of the recently proposed language representation methods.

Finally, we come to evaluate the proposed neural document language modeling framework. All the experimental results are presented at the bottom of Table 1, and several notes can be concluded from the comparisons. First, in general, the proposed framework, including three different training criteria, can deliver better results than other SOTA models by a large margin. The empirical results demonstrate the good efficacy and capacity of the proposed framework. Second, when we look into the results, a surprised finding is that the pair-wise modeling strategy is usually better than both direct and triple-wise modeling strategies. On one hand, this is because the direct modeling only takes positive examples into considerations, while the pair-wise modeling takes both relevant and irrelevant information into account. On the other hand, however, the reason why the triple-wise training strategy is worse than the pair-wise modeling should be further explored. Third, the results reveal that the performance gap between the retrieval on the manual transcripts (i.e., the TD case) and that on the recognition transcripts (i.e., the SD case) is about 6% in terms of MAP for unsupervised methods, which shows that the recognition errors inevitably mislead the statistics for both query and document so as to degrade the retrieval performance. Nevertheless, the performance gap can be largely overcome by supervised neural retrieval models, especially the proposed neural document language modeling framework. To summarize the experiments, language modeling mechanism is indeed a potential school of research to SDR, the newly language representation methods are readily to be used, and the proposed framework seems to hold practical promise for SDR and IR related applications.

**Table 1.** *Retrieval results (in MAP) achieved by various retrieval systems and the proposed framework.*

|  | TD | | SD | |
| --- | --- | --- | --- | --- |
|  | Long | Short | Long | Short |
| VSM | 0.548 | 0.339 | 0.484 | 0.273 |
| CBOW | 0.563 | 0.358 | 0.500 | 0.307 |
| Skipgram | 0.567 | **0.385** | 0.508 | **0.364** |
| GloVe | 0.558 | 0.371 | 0.502 | 0.321 |
| DM | 0.558 | 0.344 | 0.484 | 0.302 |
| DBOW | **0.579** | 0.362 | **0.540** | 0.345 |
| EV | 0.571 | 0.382 | 0.518 | **0.364** |
| QLM | 0.634 | 0.371 | 0.563 | 0.321 |
| LDA | 0.643 | 0.401 | 0.581 | 0.341 |
| GPR | 0.589 | 0.404 | 0.527 | 0.389 |
| LPEV | 0.684 | 0.418 | 0.556 | 0.390 |
| DSSM | **0.691** | 0.462 | **0.687** | **0.435** |
| Vanilla BERT | 0.617 | **0.466** | 0.568 | 0.427 |
| Direct | 0.733 | 0.454 | 0.683 | 0.396 |
| Pair-wise | **0.773** | **0.527** | **0.760** | 0.484 |
| Triple-wise | 0.738 | 0.508 | 0.728 | **0.498** |

### 5. CONCLUSION

In this paper, we have presented a neural document language modeling framework for SDR, which not only reveals a potential way to employ the pretrained language representation methods but also introduces three training criteria for deriving the model parameters. Experimental results demonstrate the remarkable superiority than other strong baselines compared in the paper, thereby indicating the potential of the enhanced document language modeling framework. For future work, we will explore the incorporation of extra cues, such as acoustic statistics and sub-word information, into the proposed framework for the SDR task. Moreover, we also plan to evaluate the framework on other downstream NLP-related tasks.

# 6. REFERENCES

[1] L.-S. Lee and B. Chen, "Spoken document understanding and organization," *IEEE Signal Processing Magazine*, 22(5):42–60, 2005.

[2] C. Chelba, T. J. Hazen and M. Saraclar, "Retrieval and browsing of spoken content," *IEEE Signal Processing Magazine*, 25(3), 2008.

[3] M. Ostendorf, "Speech technology and information access," *IEEE Signal Processing Magazine*, pp. 150–152, 2008.

[4] M. Larson and G. J. F. Jones, "Spoken content retrieval: a survey of techniques and technologies", *Foundations and Trends in Information Retrieval*, 5(4–5):235–422, 2012.

[5] I. Szoke, L. Burget, J. Cernocky and M. Fapso, "Sub-word modeling of out of vocabulary words in spoken term detection," in *Proc. of SLT*, 2008.

[6] C. Carpineto and G. Romano, "A survey of automatic query expansion in information retrieval," *ACM Comput. Surv.*, 44(1), 2012.

[7] C. D. Manning, P. Raghavan and H. Schtze, *Introduction to information retrieval*, New York: Cambridge University Press, 2008.

[8] R. A. Baeza-Yates and B. Ribeiro-Neto, *Modern information retrieval: the concepts and technology behind search*, Addison-Wesley Longman Publishing Co., Inc., 2011.

[9] Y. LeCun, Y. Bengio and G. Hinton, "Deep learning," *Nature*, 521:436–444, 2015.

[10] K. He, X. Zhang, S. Ren and J. Sun, "Deep Residual Learning for Image Recognition," in *Proc. of CVPR*, 2016.

[11] Natural Language Computer Group, Microsoft Reserach. Asia, "R-NET: machine reading comprehension with self-matching networks," Technical Report, 2017.

[12] A. Vaswani, N. Shazeer, N. Parmar, J. Uszkoreit, L. Jones, A. N Gomez, L. Kaiser, and I. Polosukhin, "Attention is all you need," in *Proc. of NIPS*, 2017.

[13] Y. Bengio, R. Ducharme, P. Vincent, and C. Jauvin, "A neural probabilistic language model," *Journal of Machine Learning Research* (3), pp. 1137–1155, 2003.

[14] T. Mikolov, K. Chen, G. Corrado, and J. Dean, "Efficient estimation of word representations in vector space," in *Proc. of ICLR*, 2013.

[15] J. Pennington, R. Socher, and C. D. Manning, "GloVe: Global vector for word representation," in *Proc. of EMNLP*, 2014.

[16] Q. Le and T. Mikolov, "Distributed representations of sentences and documents," in *Proc. of ICML*, 2014.

[17] G. Salton, A. Wong, and C. S. Yang, "A vector space model for automatic indexing," *Communications of the ACM*, 18(11), pp. 613-620, Nov. 1975.

[18] K. S. Jones, S. Walker, and S. E. Robertson. "A probabilistic model of information retrieval: development and comparative experiments (parts 1 and 2)," *Information Processing and Management*, 36(6), pp. 779-840, 2000.

[19] C. Zhai and J. Lafferty, "A study of smoothing methods for language models applied to ad hoc information retrieval," in *Proc. of SIGIR*, 2001.

[20] K.-Y. Chen, S.-H. Liu, B. Chen and H.-M. Wang, "Essence Vector-based Query Modeling for Spoken Document Retrieval," in *Proc. of ICASSP*, 2018.

[21] T. Hoffmann, "Unsupervised learning by probabilistic latent semantic analysis," *Machine Learning*, 42, pp. 177-196, 2001.

[22] D. M. Blei, A. Y. Ng and M. I. Jordan "Latent Dirichlet Allocation," in *Proc. of JMLR*, 2003.

[23] X. Wei and B. Croft, "LDA-based document models for ad-hoc retrieval," in *Proc. of SIGIR*, 2006.

[24] Y. Lv and C. X. Zhai, "Positional language models for information retrieval", in *Proc. of SIGIR*, 2009.

[25] D. Hiemstra, S. Robertson, and H. Zaragoza, "Parsimonious language models for information retrieval," in *Proc. of SIGIR*, 2004.

[26] M. E. Peters, M. Neumann, M. Iyyer, M. Gardner, C. Clark, K. Lee and L. Zettlemoyer "Deep contextualized word representations," arXiv:1802.05365, 2018.

[27] A. Radford, K. Narasimhan, T. Salimans and I. Sutskever, "Improving Language Understanding by Generative Pre-Training," OpenAI, 2018.

[28] J. Devlin, M.-W. Chen K. Lee and K. Toutanova, "BERT: Pre-training of Deep Bidirectional Transformers for Language Understanding," arXiv:1802.05365, 2018.

[29] Z. Yang, Z. Dai, Y. Yang, J. Carbonell, R. Salakhutdinov and Q. V. Le, "XLNet: Generalized Autoregressive Pretraining for Language Understanding," arXiv:1906.08237, 2019.

[30] Y. Liu, M. Ott, N. Goyal, J. Du, M. Joshi, D. Chen, O. Levy, M. Lewis, L. Zettlemoyer and V. Stoyanov, "RoBERTa: A Robustly Optimized BERT Pretraining Approach," arXiv:1907.11692, 2019.

[31] Z. Lan, M. Chen, S. Goodman, K. Gimpel, P. Sharma and R. Soricut, "A Lite BERT for Self-supervised Learning of Language Representations," arXiv:1909.11942, 2019.

[32] Y. Liu, "Fine-tune BERT for Extractive Summarization," arXiv:1903.10318, 2019.

[33] W. Yang, H. Zhang and J. Lin, "Simple Applications of BERT for Ad Hoc Document Retrieval," arXiv:1903.10972, 2019.

[34] Y. Qiao, C. Xiong, Z. Liu, and Z. Liu, "Understanding the Behaviors of BERT in Ranking," arXiv:1904.07531, 2019.

[35] J. Weston and C. Watkins, "Support Vector Machines for Multi-Class Pattern Recognition," in *Proc. of ESANN*, 1999.

[36] LDC, "Project topic detection and tracking," Linguistic Data Consortium, 2000.

[37] H. Meng, S. Khudanpur, G. Levow, D. Oard, and H. M. Wang, "Mandarin–English information (MEI): investigating translingual speech retrieval," *Computer Speech and Language*, 18(2), pp. 163–179, April 2004.

[38] J. Garofolo, G. Auzanne, and E. Voorhees, "The TREC spoken document retrieval track: A success story," in *Proc. of TREC*, 2000.

[39] K.-Y. Chen, S.-H. Liu, B. Chen, H.-M. Wang, "Essense vector-based query modeling for spoken document retrieval," in *Proc. of ICASSP*, 2018.

[40] Z.-Y. Wu, L.-P. Yen and K.-Y. Chen, "Generating Pseudo-relevant Representations for Spoken Document Retrieval," in *Proc. of ICASSP*, 2019.

[41] K.-Y. Chen, S.-H. Liu, B. Chen and H.-M. Wang, "A Locality-Preserving Essence Vector Modeling Framework for Spoken Document Retrieval," in *Proc. of ICASSP*, 2017.

[42] P.-S. Huang, X. He, J. Gao, L. Deng, A. Acero, and L. Heck, "Learning deep structured semantic models for web search using clickthrough data," in *Proc. of CIKM*, 2013.